\documentclass[conference]{IEEEtran}
\IEEEoverridecommandlockouts
\usepackage{cite}
\usepackage{amsmath,amssymb,amsfonts}
\usepackage{algorithmic}
\usepackage{multirow}
\usepackage{graphicx}
\usepackage{textcomp}
\usepackage{xcolor}
\usepackage{graphicx}
\usepackage{subcaption}
\usepackage{hyperref} 
\usepackage{booktabs}
\usepackage[table,xcdraw]{xcolor} 
\usepackage{array}
\usepackage{caption} 
\def\BibTeX{{\rm B\kern-.05em{\sc i\kern-.025em b}\kern-.08em
    T\kern-.1667em\lower.7ex\hbox{E}\kern-.125emX}}

\begin{document}

\title{\vspace*{20pt}\Large A Two-Stage Framework for Ego-Centric Key Object Identification via Object State Prediction}
\author{Shihong Ling$^{1}$, Yue Wan$^{1}$, Xiaowei Jia$^{1}$, and Na Du$^{1}$
\thanks{$^{1}$School of Computing and Information, University of Pittsburgh, Pittsburgh, PA, USA}}

\maketitle

\begin{abstract}
This paper presents a novel framework designed to enhance key object identification in autonomous driving. Existing methods primarily focus on either detecting objects independently or leveraging visual relationships, but they do not explicitly consider the ego vehicle's perspective in determining object importance. To address this gap, we propose a structured approach that integrates a virtual ego-vehicle representation and a modular object state predictor, enabling a more accurate estimation of object behaviors relative to the ego-vehicle. Subsequently, our framework employs spatial-temporal reasoning to refine key object identification, prioritizing objects based on their states and relative spatial information rather than relying solely on visual relationships. Experimental results on real-world driving datasets demonstrate the effectiveness of our approach in accurately detecting critical objects in complex traffic environments.
\end{abstract}


\section{Introduction}
In recent years, the evolution of autonomous vehicles (AVs) has witnessed remarkable technological breakthroughs, with multiple industry leaders worldwide demonstrating stable and safe driving capabilities in complex urban environments \cite{b1}. However, due to the lack of trust in AVs, drivers may perceive AVs as ``black boxes", causing unnecessary interventions \cite{b2}. Hence, it remains a critical challenge how to enhance the \textcolor{black}{transparency} of AVs' decision-making process. In other words, AVs should be able to explain the surrounding objects' states and identify the critical objects that may affect their driving as needed for the drivers.


Early methods for key object identification in driving scenarios primarily relied on convolutional neural networks (CNN)-based \cite{b37} object detection, which treats all surrounding objects as independent entities and evaluates their importance relative to the ego-vehicle \cite{b3,b4,b5,b6,b7,b8,b9}. While easy to implement, this approach does not incorporate inter-object relationships, limiting its effectiveness. To address this, some studies introduced explicit relational modeling through handcrafted annotations that demonstrate interactions among objects \cite{b10,b11,b12}. While this improves identifying the key objects, it introduces challenges related to the cost of manual annotations, the restricted variety of labeled relationships, and potential inconsistencies between human-defined relationships and machine-learned patterns. 
More recently, several works were proposed to infer the object relationship, eliminating the need for manual annotations. For example, by leveraging the self-attention mechanism, transformer-based models \cite{b38} have been utilized to capture inter-object relationships and automatically infer object interactions within the driving scene. Similar approach has been shown to further enhance key object identification performance by incorporating auxiliary tasks relevant to ego-vehicle behavior prediction \cite{b14}.
Building upon transformer-based architectures, vision-language models (VLMs) integrate both visual and textual information to enhance scene understanding \cite{b15,b16}. Although these models can provide more detailed interpretations of driving scenes, they remain limited in precisely localizing the key objects and generating relevant explanations based on key objects. 

In particular, the object-level self-attention in transformer-based models is focused on capturing interactions between all objects, but they are not designed to distinguish between relevant and irrelevant interactions. As a result, the spurious relationship between irrelevant objects may add additional noise in driving scene understanding. In contrast, the interactions between surrounding objects and the ego-vehicle should be highlighted. 
To make the interaction more ego-vehicle-centric, recent methods \cite{b13, b14} incorporate the state of the ego-vehicle (e.g., velocity, planned direction and action) as an additional element into the computation of self-attention. 
Despite its emphasis on ego-vehicle-centric interaction, an important aspect of driving involves reacting to external changes, rather than solely following a predefined ego-vehicle action plan. For example, when a vehicle abruptly cuts in, the ego-vehicle must react accordingly, regardless of its initial planned maneuver. This highlights the need for models to dynamically infer and adapt to the concurrent states of surrounding objects, in addition to considering the state of the ego-vehicle (e.g., intended actions). 

To address these challenges, we propose a two-stage framework that enhances key object identification in autonomous driving. In the first stage, the \textit{Object State Predictor} infers the states of surrounding objects based on their spatial-temporal information with respect to the ego-vehicle. In the second stage, the \textit{Key Object Identifier} determines the most critical object in the scene by leveraging the predicted object states and their relevance to the ego-vehicle. One key innovation of our framework is the incorporation of dynamic object states that are relevant to the ego-vehicle rather than relying solely on visual relationships between objects. We present a virtual representation for the ego-vehicle to provide a consistent spatial reference for estimating the object states. 
To further enhance the accuracy of object state prediction, we employ a modular design, where separate models are trained for different object categories instead of using a single unified model. We observe from experiments that it improves accuracy by allowing the model to specialize in distinct object types. Through evaluations on our diverse driving scenario datasets, we validate that our framework significantly improves key object identification. 

\section{Related Work}

\subsection{Methods for Key Object Identification}
Traditional methods treat key object identification as a straightforward object detection problem, where all detected objects are considered independently \cite{b3,b4,b5,b6,b7,b8,b9}. More advanced methods introduce relational modeling, incorporating object interactions to improve key object identification \cite{b10,b11,b12}. Among recent advancements, transformer-based models have demonstrated superior performance due to their ability to capture long-range dependencies and model relationships among objects \cite{b13, b14, b15, b16}. For instance, DRUformer \cite{b13} integrates a self-attention mechanism to infer object importance based on their interactions within the scene. Another approach formulates object importance estimation as a binary classification problem, leveraging relational reasoning and self-supervised learning to reduce reliance on manually annotated relationships \cite{b14}. These methods enhance object importance estimation without requiring explicit relationship annotations, making them more scalable.

Inspired by these transformer-based architectures, our method employs an LSTM-based encoder to capture temporal dependencies in object states, followed by a transformer module for key object identification. To explicitly incorporate the ego-vehicle's influence, we introduce a virtual ego-vehicle representation, which provides a consistent spatial reference for evaluating object importance. Unlike previous methods that rely solely on inter-object relationships, our approach selects the key object by considering the interactions between objects and the ego-vehicle. Additionally, we introduce an object state predictor that classifies object behaviors based on spatial and visual cues to support the key object identification process.

\subsection{Feature Representation for Object State Prediction}
Recent studies on behavior prediction in driving scenarios emphasize the importance of utilizing temporal-spatial and visual features \cite{b17, b18, b19, b20}. For instance, Huang et al. \cite{b18} predict pedestrian actions by extracting spatial features from ego-vehicle camera observations and computing motion features (velocity and trajectory) based on these spatial cues. Similarly, Hayakawa and Dariush \cite{b20} estimate surrounding vehicle states by deriving spatial features (2D bounding box positions) using monocular object detection and computing motion features (velocity and orientation) through depth estimation and optical flow.

To collect such features, instead of relying on multiple sensors, leveraging advanced computer vision techniques presents a cost-effective alternative. One method is to directly use the 3D object detection \cite{b21, b22}, however, these methods suffer from ground plane estimation errors which may lead to depth prediction inaccuracies. An alternative approach combines state-of-the-art 2D object detection \cite{b23,b24,b25} with depth estimation models \cite{b26,b27,b28}. Additionally, we propose using image occupancy area changes over time to infer depth by tracking object size variations across frames, reducing reliance on explicit 3D reconstruction.

\begin{figure*}[htbp]
    \centering
    \includegraphics[width=0.8\textwidth]{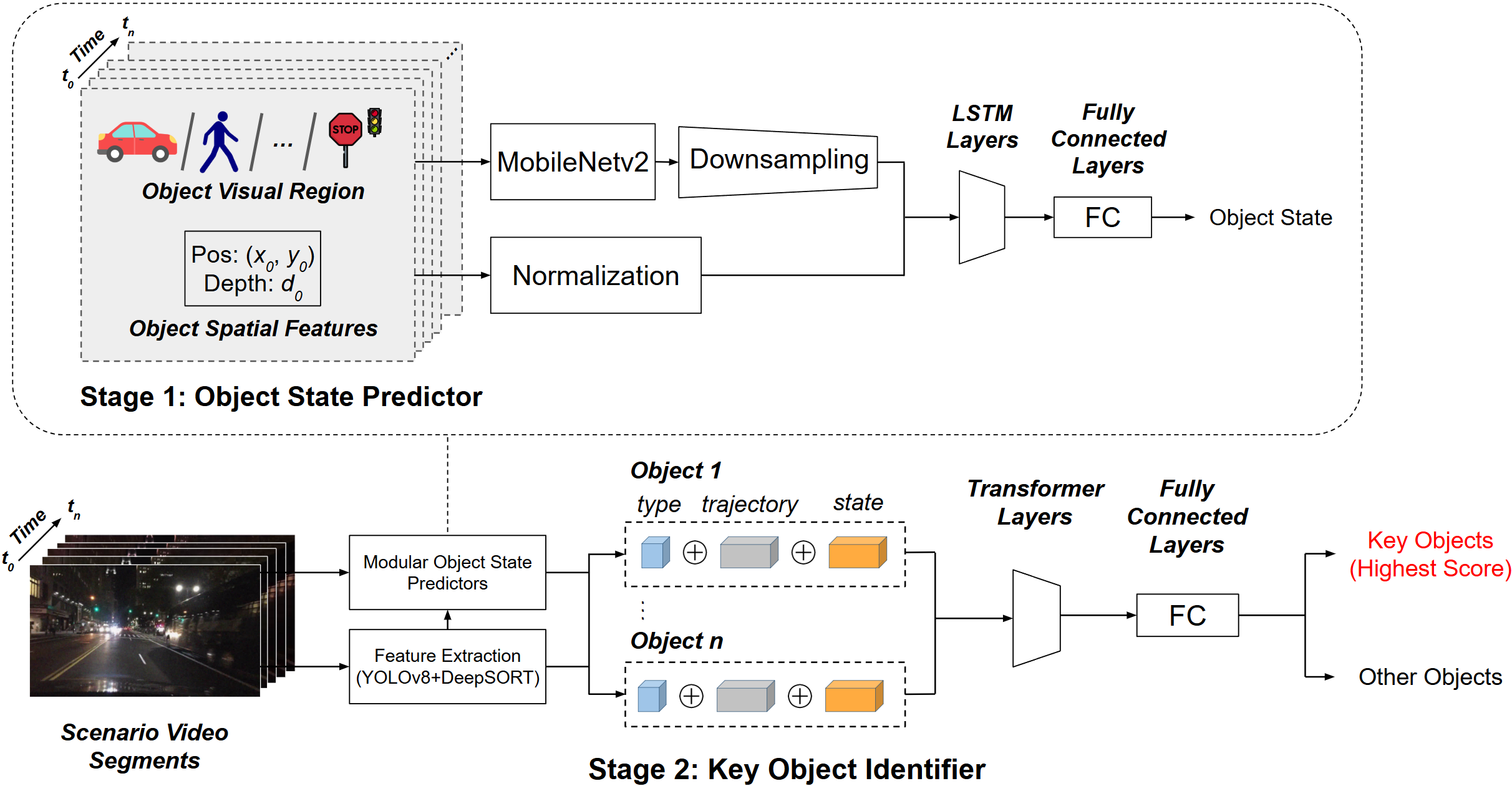}
    \caption{Two-stage model architecture: (1) Object State Predictor, which classifies object states based on spatial and visual features, and (2) Key Object Identifier, which determines the most critical object affecting the ego-vehicle.}
    \label{fig:model_architectures}
\end{figure*}
\vspace{-0.8em}

\section{Methodology}

Our proposed method is designed to identify key objects that can affect the ego-vehicle's driving. As shown in Figure~\ref{fig:model_architectures}, \textcolor{black}{the framework consists of a preprocessing stage followed by two learning stages. Prior to Stage 1, each video segment is processed by YOLOv8 \cite{b34} and DeepSORT \cite{b39} to detect and track objects, producing object trajectories and cropped object regions. In Stage 1, the Object State Predictor classifies driving-relevant object states from spatial and visual features. In Stage 2, the Key Object Identifier ranks object significance using the predicted object states together with temporal change features. The state predictor and identifier are trained separately.}

\subsection{Object State Predictor}
The \textit{Object State Predictor} takes as input a sequence of objects' spatial and visual information over time and outputs predictions of their states. Dynamic objects, such as pedestrians and vehicles, are classified into states like cut-in, block, and no impact, while static objects, including traffic lights and stop signs, are categorized separately, with traffic lights labeled as red, turning red, or no impact, and stop signs as impact or no impact. Given the distinct characteristics of different object types, we propose a modular model setup, where separate models are trained for each type. We validate these choices through feature ablations, including spatial-only, visual-only, and their combination, as well as comparing the modular models with the unified model. Details of this process are provided in the \textit{Experiment Setup} section.

To effectively capture object behaviors over time, we extract features from a 2-second video segment by sampling frames at intervals of 0.5 seconds \cite{b33, b36}. The extracted features include spatial features and visual features. 

The spatial features consist of an object's 2D position \textcolor{black}{in the image frame} and its estimated depth. For 2D position, we propose using the relative position with respect to the ego-vehicle. Since the ego-vehicle is not explicitly present in the video frames, we assume its position at the bottom-middle of the frame. This assumption allows for a consistent interpretation of object behaviors that depend on their relation to the ego-vehicle. We also experiment with the absolute position as a comparison, which records object coordinates directly within the frame. 

The relative representation provides more reliable performance. 
For estimating the depth, we explore two alternatives: 
\begin{itemize}
    \item We utilize a pre-trained depth estimation model (Intel-ISL MiDaS v3.1 model \cite{b26}, with the \texttt{dpt\_swin2\_large\_384} variant for an optimal balance between accuracy and efficiency), which infers depth from pixel values. 
    \item  We leverage the object size, which refers to the area the object occupies within the image frame. The larger area suggests proximity to the camera. 
\end{itemize}

Consequently, the spatial feature of an object in a time sequence is represented as follows:

\begin{equation}
F_{spatial} = \{ (x_t, y_t), d_t \} \quad \forall t \in [t_1, t_n]
\end{equation}
where \( (x_t, y_t) \) represents either the absolute or relative positions \textcolor{black}{in the image frame}, \( d_t \) is the estimated depth obtained either through the pre-trained MiDaS model or inferred from the object's image coverage, 
and \( t \) corresponds to the sampled timestamps extracted at 0.5-second intervals over a 2-second duration (i.e, n=5). The visual features are extracted based on the detected bounding box of the object, which defines the corresponding region in the image. This region is processed using a pre-trained MobileNetV2 \cite{b35} model to derive its feature representation. The extracted representation is then passed through a projection layer to reduce its dimensionality, aligning it with that of the spatial features. 

The final input to the model consists of concatenated relative spatial and visual features, which are fed into a \textcolor{black}{two-layer LSTM-based network with 128 and 64 hidden units, respectively, to capture temporal dependencies prior to classification.} The model is trained using the categorical cross-entropy loss to optimize multi-class predictions.

\subsection{Key Object Identifier}

The \textit{Key Object Identifier} plays a crucial role in determining which objects in a video frame should be considered most relevant to the ego-vehicle’s decision-making. This module is based on the output of the \textit{Object State Predictor}. For each detected object \( o_i \) within a video segment, we extract several key features: the object’s predicted state \( S_i \), its object type \( C_i \), and its change in position along $x$ and $y$ dimensions, and change in depth over a 2-second period. These features are combined into a comprehensive feature vector \( F_i \) for each object:

\begin{equation}
F_i = [S_i, C_i, \Delta x_i, \Delta y_i, \Delta d_i]
\end{equation}
where \( \Delta x_i, \Delta y_i, \Delta d_i \) represent the change in position and depth over the temporal window.
\textcolor{black}{These feature vectors are processed by a two-layer transformer encoder, where self-attention captures pairwise relations among the object-level feature vectors within the scene. The resulting representations are then passed through a fully connected scoring layer to generate an importance score for each object. The object with the highest score is identified as the most relevant object influencing the ego-vehicle’s decision. To optimize the importance ranking,} we apply a binary cross-entropy (BCE) loss:

\begin{equation}
L = \text{BinaryCrossEntropy}(z_{\text{pred}}, z_{\text{GT}})
\end{equation}
where \( z_{\text{pred}} \) is the predicted significance score of the object and \( z_{\text{GT}} \) is the ground truth label indicating whether the object is the most important in the scene.

\section{Experiment Setup}
Our experimental objective is twofold: (i) to select the most effective \textit{Object State Predictor} and (ii) to train the \textit{Key Object Identifier} based on the chosen predictor. To determine the optimal predictor, we explore two design dimensions: feature fusion and training regime. For feature fusion, we evaluate the following input variants: \textit{visual-only}, \textit{spatial (absolute position + estimated depth)}, \textit{spatial (relative position + estimated depth)}, \textit{spatial (absolute position + object size)}, \textit{spatial (relative position + object size)}, \textit{visual + spatial (absolute position + estimated depth)}, \textit{visual + spatial (relative position + estimated depth)}, \textit{visual + spatial (absolute position + object size)}, \textit{visual + spatial (relative position + object size)}, and \textit{visual + spatial (absolute)}. For the training regime, we develop \textit{modular} (category-specific) models and an \textit{all-in-one} baseline. All variants adhere to the same protocol, including identical data splits, optimizer/schedule, and hyperparameters; only the input representation and modularization differ. We compare state-prediction accuracy and F1 scores across variants and regimes on the validation set to identify the best-performing configuration. Unless otherwise specified, the \textit{Key Object Identifier} is subsequently trained on the outputs of the selected predictor using a fixed methodology, and its downstream key-object accuracy is reported separately.

\subsection{Experiment Dataset}
Our experiment dataset consists of 1,500 videos with human-annotated key objects, selected from BDD100K \cite{b29}, Argoverse 2 \cite{b31}, and Rank2Tell \cite{b30} for further processing. \textcolor{black}{We selected clips to cover driving-relevant object categories while preserving diversity across traffic conditions and viewpoints. In this study, the key object is defined as the single object in a 2-second segment that most directly explains the ego-vehicle’s immediate driving response under the visible scene context.} Each key object is annotated with bounding boxes, from which we derive absolute position (center of the bounding box), relative position (offset from the ego-vehicle’s assumed bottom-center position), and size (bounding box area). These features are normalized using MinMax scaling while preserving directional information. To ensure a balanced dataset, we introduce negative samples by identifying non-key objects. \textcolor{black}{Specifically, all tracked objects other than the annotated key object can be treated as non-key candidates. We utilize YOLOv8 \cite{b34} in conjunction with DeepSORT \cite{b39} to detect and track objects, employing an Intersection-over-Union (IoU) threshold of 0.3 solely as a spatial matching criterion to exclude overlapping bounding boxes and to sample non-key contextual objects surrounding the key object. The sampled non-key objects are assigned a state of \textit{no impact}.
The final dataset consists of 9,685 traffic light samples (3,228 with no impact), 1,160 stop signs (580 with no impact), 1,565 pedestrians (522 with no impact), and 4,936 vehicles (1,645 with no impact).}

\subsection{Training Procedure}

\subsubsection{Object State Predictor Training}
\textit{The Object State Predictor} is trained to classify object states over time. All feature variants listed above share identical data splits and optimization; only the input representation differs, as defined in the \textit{Methodology} section. The training process involves tuning a sequential model using a combination of spatial and visual information to predict the correct state for each object category. Given the distinct behavioral characteristics of different object types, separate models are trained for each category, ensuring optimized learning for each.

To optimize model performance, we use the Adam optimizer with a learning rate of \(1 \times 10^{-4}\) and a batch size of 32. The model undergoes training for 100 epochs with early stopping applied, monitoring validation loss with a patience of 10 epochs to prevent overfitting. For hyperparameter tuning, we employ a Random Search strategy with a maximum of 20 trials, where each trial is executed twice. The best-performing model is selected based on validation accuracy.

The loss function used for training is categorical cross-entropy, given that the task involves multi-class classification of object states. The training objective is to minimize the difference between the predicted state labels and the ground truth annotations, ensuring accurate recognition of dynamic behaviors and static states within driving scenarios.

\subsubsection{Key Object Identifier Training}
The \textit{Key Object Identifier} is trained to determine the most significant object within a scene that influences the ego-vehicle’s driving decision. Unless specified otherwise, the identifier is trained using the optimal configuration of the \textbf{Object State Predictor}. The model takes as inputs the feature representations, which are the predicted states from the Object State Predictor, of all detected objects and assigns significance scores to facilitate the identification of the most relevant object.

To establish the relationship between object significance and the ego-vehicle’s response, the model is trained to predict the key object using attention-based weighting mechanisms. The training process follows a similar optimization strategy as the \textit{Object State Predictor}, using the Adam optimizer with a learning rate of \(1 \times 10^{-4}\) and a batch size of 16. The training is performed for 100 epochs with early stopping monitoring validation loss to enhance generalization.

The objective function consists of a binary cross-entropy loss, where the model learns to distinguish between key and non-key objects based on their contextual relevance to the driving decision. The training process ensures that the model prioritizes critical elements in the scene, allowing the system to effectively recognize and highlight the most important object influencing the ego-vehicle’s behavior.

\subsubsection{Evaluation Metrics}
We evaluate the performance of our models using standard classification metrics, specifically Accuracy and F1-score. For the \textit{Object State Predictor}, accuracy measures the proportion of correctly classified object states among all predictions, while the F1-score accounts for both precision and recall, ensuring robustness to class imbalances. For the \textit{Key Object Identifier}, accuracy is used to evaluate the proportion of cases where the highest-scoring object matches the ground truth key object. With object detection, tracking, and depth estimation precomputed, the state predictor and key-object identifier run on per-object features at $\leq$0.10\, ms/frame.

\section{Results}

\subsection{Key Object Identifier Performance}
As the primary component in our framework, the \textit{Key Object Identifier} is evaluated based on its ability to correctly identify the most significant object in a given driving We compare our model's performance against two existing approaches: (1) the explainable object-induced model \cite{b32}, a recent CNN-based approach which prioritizes object relevance in scene interpretation, and (2) DRUformer \cite{b13}, one of the latest transformer-based approaches designed to enhance key object identification. For the CNN-based model, we fine-tuned it on our dataset using the last frame of each video as input, modifying its output to treat key object identification as the main task while retaining object explanation and ego-vehicle action prediction as auxiliary tasks. For DRUformer, since it was originally trained on the DRAMA dataset, which differs from our dataset in terms of frame sampling rate, input sequence length, and scene duration, we adjusted its input preprocessing pipeline accordingly and retrained it. 

Table \ref{tab:key_object_results} presents the performance comparison. Our proposed method outperforms both baselines upon the support of the best-performing Object State Predictor, demonstrating its ability to more accurately identify key objects across different categories. The results show that stop signs achieve the highest accuracy due to their static nature and consistent positioning, whereas dynamic objects like cars and pedestrians pose greater challenges due to variations in movement and interactions.
\newcommand{\std}[1]{{\scriptsize$\pm$#1}}
\begin{table}[htbp]
  \centering
  \caption{Key Object Identifier accuracy by key-object category.}
  \label{tab:key_object_results}
  \vspace{0.2em}
  \small
  \setlength{\tabcolsep}{6pt}
  \renewcommand{\arraystretch}{1.06}
  \begin{tabular*}{\linewidth}{@{\extracolsep{\fill}} l c c c}
    \toprule
    \textbf{Object} & \textbf{Ours} & \textbf{Object-induced Model} & \textbf{DRUformer} \\
    \midrule
    Car           & \textbf{0.78\std{.02}} & 0.62\std{.01} & 0.69\std{.02} \\
    Person        & \textbf{0.76\std{.02}} & 0.65\std{.02} & 0.71\std{.01} \\
    Traffic Light & \textbf{0.71\std{.01}} & 0.59\std{.01} & 0.63\std{.01} \\
    Stop Sign     & \textbf{0.94\std{.01}} & 0.71\std{.02} & 0.82\std{.01} \\
    \bottomrule
  \end{tabular*}
\end{table}
\vspace{-0.8em}
\subsection{Object State Predictor Performance}
Since the \textit{Key Object Identifier} depends on the \textit{Object State Predictor}, we evaluate a comprehensive set of feature variants to identify the most effective predictor as introduced in \textit{Experiment Setup} section. These model variants follow a modular structure where separate networks are trained for different object categories, allowing optimal adaptation to category-specific patterns.

Table \ref{tab:modular_results} presents the results for the modular models. The highest accuracy and F1 score for each feature combination are highlighted in bold. The results indicate that using relative position representation generally outperforms absolute position in most cases, particularly for traffic lights and stop signs. Additionally, object size-based depth estimation proves to be more reliable for static objects, while pre-trained depth estimation is more effective for dynamic objects.

\begin{table}[htbp]
  \centering
  \caption{Modular Object State Predictor performance across feature combinations (Accuracy$\mid$F1, mean$\pm$std). Abbrev.: \textbf{vis}=visual, \textbf{sp}=spatial, \textbf{abs}=absolute position, \textbf{rel}=relative position, \textbf{dep}=estimated depth (MiDaS), \textbf{size}=object-size depth proxy.}
  \label{tab:modular_results}
  \vspace{0.2em}
  \small
  \setlength{\tabcolsep}{5pt}
  \renewcommand{\arraystretch}{1.06}

  \begin{tabular*}{\linewidth}{@{\extracolsep{\fill}} l l c}
    \toprule
    \textbf{Pred. Type} & \textbf{Feat. Comb.} & \textbf{Perf. (Accuracy$\mid$F1)} \\
    \midrule

    \multirow{10}{*}{Pedestrian}
      & vis-only                    & 0.54\std{.03}$\mid$0.50\std{.03} \\
      & sp(abs+dep)                 & 0.58\std{.03}$\mid$0.53\std{.03} \\
      & sp(rel+dep)                 & 0.66\std{.02}$\mid$0.62\std{.02} \\
      & sp(abs+size)                & 0.78\std{.02}$\mid$0.75\std{.02} \\
      & sp(rel+size)                & 0.81\std{.02}$\mid$0.76\std{.02} \\
      & vis+sp(abs+dep)             & 0.60\std{.03}$\mid$0.55\std{.03} \\
      & vis+sp(rel+dep)             & 0.68\std{.02}$\mid$0.64\std{.02} \\
      & vis+sp(abs+size)            & 0.80\std{.02}$\mid$0.77\std{.02} \\
      & \textbf{vis+sp(rel+size)}   & \textbf{0.83\std{.01}} $\mid$ \textbf{0.78\std{.02}} \\
      & vis+sp(abs)                 & 0.79\std{.02}$\mid$0.76\std{.02} \\
    \midrule

    \multirow{10}{*}{Car}
      & vis-only                    & 0.58\std{.03}$\mid$0.56\std{.03} \\
      & sp(abs+dep)                 & 0.60\std{.03}$\mid$0.58\std{.03} \\
      & sp(rel+dep)                 & 0.63\std{.02}$\mid$0.61\std{.02} \\
      & sp(abs+size)                & 0.73\std{.02}$\mid$0.72\std{.02} \\
      & sp(rel+size)                & 0.74\std{.02}$\mid$0.73\std{.02} \\
      & vis+sp(abs+dep)             & 0.62\std{.03}$\mid$0.60\std{.03} \\
      & vis+sp(rel+dep)             & 0.65\std{.02}$\mid$0.63\std{.02} \\
      & vis+sp(abs+size)            & 0.75\std{.02}$\mid$0.74\std{.02} \\
      & \textbf{vis+sp(rel+size)}   & \textbf{0.77\std{.02}} $\mid$ \textbf{0.76\std{.02}} \\
      & vis+sp(abs)                 & 0.74\std{.02}$\mid$0.73\std{.02} \\
    \midrule

    \multirow{10}{*}{Traffic Light}
      & vis-only                    & 0.88\std{.01}$\mid$0.85\std{.01} \\
      & sp(abs+dep)                 & 0.78\std{.02}$\mid$0.73\std{.02} \\
      & sp(rel+dep)                 & 0.83\std{.02}$\mid$0.79\std{.02} \\
      & sp(abs+size)                & 0.90\std{.01}$\mid$0.87\std{.01} \\
      & sp(rel+size)                & 0.90\std{.01}$\mid$0.86\std{.01} \\
      & vis+sp(abs+dep)             & 0.80\std{.02}$\mid$0.75\std{.02} \\
      & vis+sp(rel+dep)             & 0.85\std{.01}$\mid$0.81\std{.02} \\
      & \textbf{vis+sp(abs+size)}   & \textbf{0.92\std{.01}} $\mid$ \textbf{0.89\std{.01}} \\
      & vis+sp(rel+size)            & 0.92\std{.01}$\mid$0.88\std{.01} \\
      & vis+sp(abs)                 & 0.90\std{.01}$\mid$0.86\std{.01} \\
    \midrule

    \multirow{10}{*}{Stop Sign}
      & vis-only                    & 0.97\std{.01}$\mid$0.95\std{.01} \\
      & sp(abs+dep)                 & 0.84\std{.02}$\mid$0.82\std{.02} \\
      & sp(rel+dep)                 & 0.90\std{.01}$\mid$0.88\std{.01} \\
      & sp(abs+size)                & 0.97\std{.01}$\mid$0.95\std{.01} \\
      & sp(rel+size)                & 0.97\std{.01}$\mid$0.95\std{.01} \\
      & vis+sp(abs+dep)             & 0.85\std{.02}$\mid$0.83\std{.02} \\
      & vis+sp(rel+dep)             & 0.91\std{.01}$\mid$0.89\std{.01} \\
      & \textbf{vis+sp(abs+size)}   & \textbf{0.98\std{.00}} $\mid$ \textbf{0.96\std{.01}} \\
      & vis+sp(rel+size)            & 0.98\std{.01}$\mid$0.96\std{.01} \\
      & vis+sp(abs)                 & 0.97\std{.01}$\mid$0.95\std{.01} \\
    \bottomrule
  \end{tabular*}
\end{table}
We also evaluate the performance of an all-in-one model, where a single network handles all object categories. Similar to the modular approach, we experiment with different feature combinations. Table \ref{tab:all_in_one_results} presents the best-performing configuration for the all-in-one model. While this configuration achieves competitive results, it still underperforms compared to the best modular models, particularly for objects with distinct behavior patterns such as pedestrians and cars. 

\begin{table}[htbp]
  \centering
  \caption{All-in-one vs. modular-best performance by object category. Best modular configurations: Ped/Car = \textit{vis+sp(rel+size)}; TL/SS = \textit{vis+sp(abs+size)} (see Table~\ref{tab:modular_results} for abbreviations).}
  \label{tab:all_in_one_results}
  \vspace{0.2em}
  \small
  \setlength{\tabcolsep}{6pt}
  \renewcommand{\arraystretch}{1.06}
  \begin{tabular*}{\linewidth}{@{\extracolsep{\fill}} l c c}
    \toprule
    \textbf{Object Type} & \textbf{All-in-one (ACC$\mid$F1)} & \textbf{Modular-best (ACC$\mid$F1)} \\
    \midrule
    Car           & 0.64\std{.02}$\mid$0.62\std{.02} & \textbf{0.77\std{.02}$\mid$0.76\std{.02}} \\
    Pedestrian    & 0.71\std{.01}$\mid$0.67\std{.02} & \textbf{0.83\std{.01}$\mid$0.78\std{.02}} \\
    Traffic Light & 0.78\std{.01}$\mid$0.74\std{.01} & \textbf{0.92\std{.01}$\mid$0.89\std{.01}} \\
    Stop Sign     & 0.85\std{.01}$\mid$0.80\std{.01} & \textbf{0.98\std{.00}$\mid$0.96\std{.01}} \\
    \bottomrule
  \end{tabular*}
\end{table}

\subsection{Visualized Improvements \& Limitations}
\begin{figure*}[htbp]
    \centering
    \includegraphics[width=1.0\textwidth]{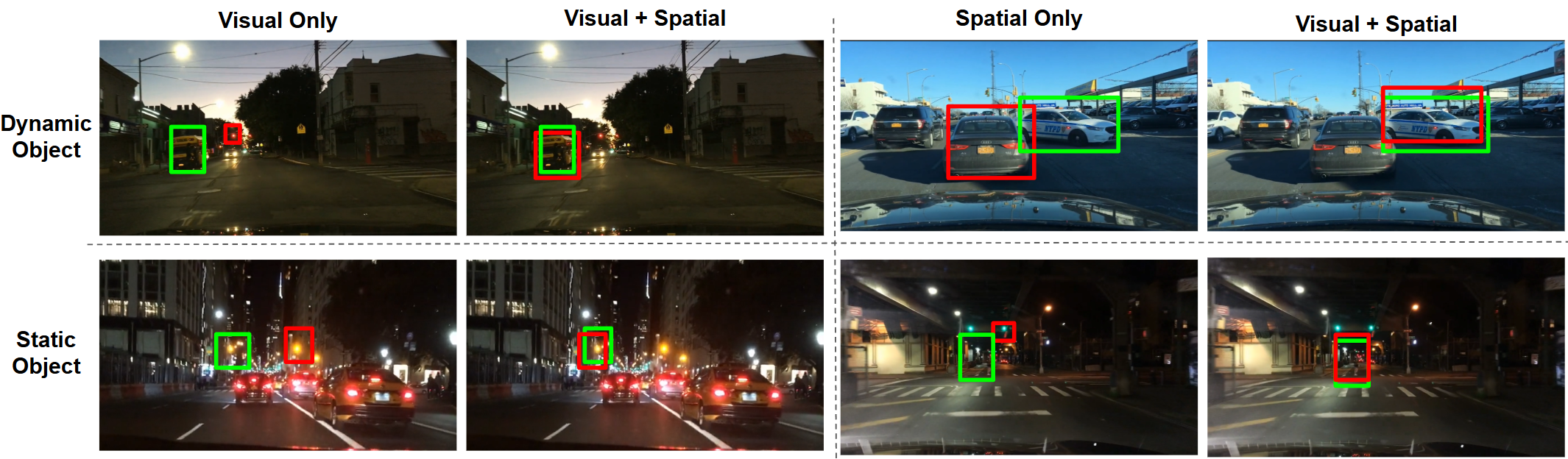} 
    \caption{Qualitative comparison across feature configurations. Rows: \textbf{dynamic} (top) and \textbf{static} (bottom) scenes. Columns show \textit{Visual-only}, \textit{Visual+Spatial}, \textit{Spatial-only}, and \textit{Visual+Spatial}. \textcolor{green}{Green} boxes denote ground-truth key objects; \textcolor{red}{red} boxes denote predictions. Each panel uses the strongest model under the corresponding condition (selected on validation Accuracy/F1).}
    \label{fig:qual_improve}
\end{figure*}

\begin{figure*}[htbp]
    \centering
    \includegraphics[width=1.0\textwidth]{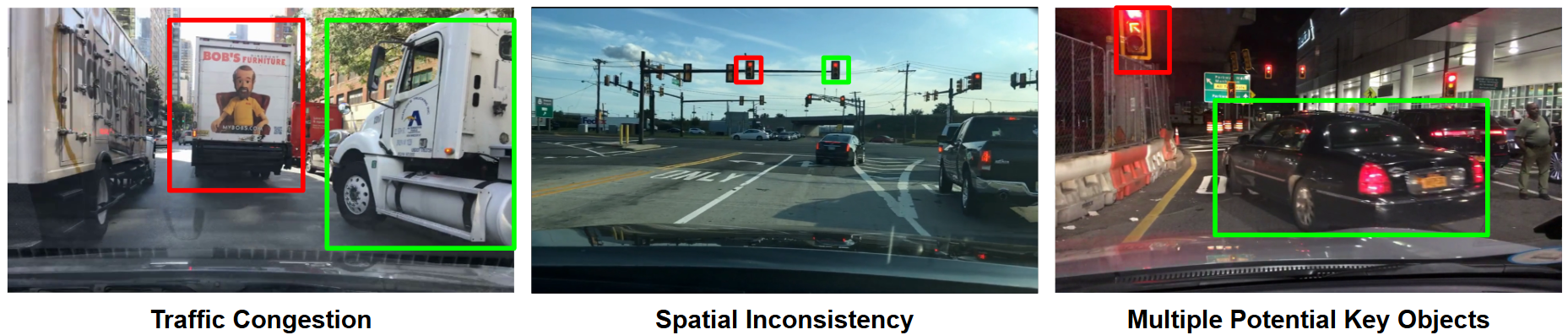}
    \caption{Representative failure cases. Left: \textbf{Traffic congestion}. Middle: \textbf{Spatial inconsistency} across camera viewpoints. Right: \textbf{Multiple potential key objects}. \textcolor{green}{Green} boxes mark ground-truth key objects; \textcolor{red}{red} boxes show model predictions.}
    \label{fig:qual_limit}
\end{figure*}

Figure~\ref{fig:qual_improve} provides several case studies, complementing the quantitative results, in how the fusion of spatial cues with visual appearance enhances key-object identification in both dynamic and static scenes, using the optimal model for each condition. In dynamic scenarios (top row), the \textit{Visual-only} approach incorrectly selects a traffic light, whereas the \textit{Visual+Spatial} approach accurately identifies the crossing pedestrian. Similarly, \textit{Spatial-only} focuses on the stopped car ahead, while \textit{Visual+Spatial} selects the crossing police vehicle. Although our dataset does not explicitly differentiate between emergency and regular vehicles, appearance cues help clarify relevance when combined with spatial context. In static scenarios (bottom row), \textit{Visual-only} highlights a signal from an adjacent approach, but adding spatial cues shifts attention to the lane-consistent light. Conversely, \textit{Spatial-only} misinterprets a green light as the key object, whereas \textit{Visual+Spatial} correctly prioritizes the pedestrian.

However, challenges persist for some scenes (Fig.~\ref{fig:qual_limit}). In \textit{traffic congestion}, the model may label nearby but non-influential items as key objects when multiple movers compete for attention. Under \textit{spatial inconsistency} caused by varying camera placements, static objects, such as traffic lights from adjacent approaches, are occasionally attributed to the ego lane, resulting in lane-mismatched selections. In scenes with \textit{multiple potential key objects}, discerning the truly causal one requires more advanced reasoning. For example, when a lead vehicle stops while leaving a buffer before the stop line, selecting the red signal (prediction) is suboptimal—the stopped vehicle (ground truth) is more informative because its behavior implies an occluded, unexpected event ahead.

\section{Discussion}
By leveraging the most effective \textit{Object State Predictor}, \textit{Key Object Identifier} achieves strong performance in determining the most significant object in a scene. Our approach surpasses both the CNN-based model adapted from the explainable object induced model \cite{b10} and the transformer-based model adapted from DRUformer \cite{b11}, demonstrating that explicitly modeling object states and their spatial relationships with the ego-vehicle improves key object detection.

In addition, by comparing feature representations and model architectures (Tables~\ref{tab:modular_results} and \ref{tab:all_in_one_results}), the \textit{modular} Object State Predictor—trained per category—consistently surpasses the \textit{all-in-one} model across all classes: Pedestrian (+0.12 Accuracy / +0.11 F1; 0.83/0.78 vs. 0.71/0.67), Car (+0.13/+0.14; 0.77/0.76 vs. 0.64/0.62), Traffic Light (+0.14/+0.15; 0.92/0.89 vs. 0.78/0.74), and Stop Sign (+0.13/+0.16; 0.98/0.96 vs. 0.85/0.80). Experiments over different variants show that (1) for dynamic classes (Pedestrian, Car), \emph{relative} position outperforms \emph{absolute} and the object-size depth proxy beats estimated depth in spatial-only settings; fusing visual with the best spatial features yields small but consistent gains (e.g., Pedestrian: 0.83/0.78 vs. 0.81/0.76, +0.02/+0.02; Car: 0.77/0.76 vs. 0.74/0.73, +0.03/+0.03). (2) For static classes (Traffic Light, Stop Sign), spatial features with \emph{size} dominate depth-based ones (e.g., Traffic Light: 0.90–0.92 with size vs. 0.78–0.85 with depth; Stop Sign: 0.97–0.98 vs. 0.85–0.91), and adding visual features provides only modest improvements (Traffic Light: +0.02 Accuracy/+0.02 F1; Stop Sign: +0.01/+0.01). Finally, the \textit{Key Object Identifier}, trained on the best state predictor, attains the highest accuracy among baselines for all key-object categories (Table~\ref{tab:key_object_results}), with gains over DRUformer of +0.09 (Car), +0.05 (Person), +0.08 (Traffic Light), and +0.12 (Stop Sign).

These quantitative findings are reflected in the qualitative examples shown in Figure~\ref{fig:qual_improve}. In dynamic scenes, the \textit{Visual-only} approach is easily distracted by salient but behaviorally irrelevant cues, such as selecting a red light, while the \textit{Spatial-only} approach often prioritizes the nearest object, like a stopped lead car. Combining these approaches highlights the truly causal agents, such as a crossing pedestrian or a police vehicle, which aligns with the improvements reported for Pedestrian and Car. In static scenes, spatial cues align predictions with the ego lane and distance, helping to disambiguate multiple signals. Meanwhile, visual evidence prevents the misinterpretation of a green light as an actual hazard, such as a nearby pedestrian. These examples collectively illustrate the complementary roles of spatial geometry and visual appearance.


Despite the strong performance demonstrated by both quantitative and qualitative results, certain limitations persist (Fig.~\ref{fig:qual_limit}). Firstly, in high-density scenes, such as heavy traffic or crowded crossings, the model may over-select nearby but non-influential agents, thereby reducing key-object precision. This issue is probably caused by the short temporal window of 2 seconds at 0.5-second sampling, which inadequately represents intent, and the reliance on proximity-biased signals without explicit interaction reasoning. A focused remedy involves extending the temporal context and applying temporal attention over 4 to 6 seconds to ensure that intent and interaction cues suppress near-but-non-causal agents. Secondly, cross-dataset camera variations introduce spatial inconsistency, occasionally misassigning signals from adjacent approaches to the ego lane. This is largely due to viewpoint differences that misalign the virtual ego reference at bottom-center and monocular depth scale drift across cameras. A practical solution is camera-aware spatial normalization via homography/IPM to a lane-anchored bird’s-eye view, which stabilizes geometry across viewpoints. Thirdly, when multiple plausible key objects coexist, selecting the truly causal one necessitates deeper reasoning. For example, a lead vehicle stopping before the line may indicate a hidden hazard and be more critical than the red light itself. The likely gaps include the absence of an explicit causal objective and reliance on instantaneous salience. An effective remedy may be further incorporating a causal reasoning head trained with temporal supervision to rank objects by inferred influence on the ego vehicle action.

\section{Future Work}
Future work will address the three limitations identified above. For high-density scenes, we will extend the temporal window (e.g., 4–6 seconds) and incorporate temporal attention and lightweight interaction reasoning to down-weight near-but-non-causal agents. Additionally, we will implement hard-negative mining for crowded scenarios. To mitigate spatial inconsistency across cameras, we plan to use camera-aware spatial normalization, lane/topology priors, and scale-consistent depth, with optional sensor fusion when available. To better resolve cases with multiple plausible key objects, we will introduce a causal reasoning head trained with temporal supervision to rank objects by their influence on the ego action and report the top-k candidates with brief rationales. Beyond modeling, we will expand the detection taxonomy (e.g., construction signs, emergent vehicles, cyclists) and curate failure-focused data to improve robustness.

\section{Conclusion}
We propose a framework to enhance key object identification by understanding the states of surrounding objects with respect to the ego-vehicle. By introducing a virtual ego-vehicle representation, our model ensures a consistent spatial reference for improved object state estimation. A modular object state predictor further enhances classification accuracy, while spatial-temporal analysis prioritizes objects based on dynamic interactions rather than solely visual cues. Experiments show our framework outperforms baselines in identifying critical objects and improving scenario understanding. However, challenges remain in handling dense traffic and rare driving scenarios. Future work will focus on expanding dataset diversity and improving model generalization.



\begin{thebibliography}{50}
\bibitem{b1} G. Lan and Q. Hao, ``End-to-end Planning of Autonomous Driving in Industry and Academia: 2022-2023,'' 2023, doi: 10.48550/arXiv.2401.08658.

\bibitem{b2} C. Rudin, ``Stop Explaining Black Box Machine Learning Models for High-stakes Decisions and Use Interpretable Models Instead,'' \textit{Nature Machine Intelligence}, vol. 1, pp. 206--215, 2019, doi: 10.1038/s42256-019-0048-x.

\bibitem{b3} S. Alletto, A. Palazzi, F. Solera, S. Calderara, and R. Cucchiara, ``Dr(eye)ve: A Dataset for Attention-Based Tasks with Applications to Autonomous and Assisted Driving,'' in \textit{2016 IEEE Conference on Computer Vision and Pattern Recognition Workshops (CVPRW)}, Jun 2016, doi: 10.1109/CVPRW.2016.14.

\bibitem{b4} A. Tawari, P. Mallela, and S. Martin, ``Learning to Attend to Salient Targets in Driving Videos Using Fully Convolutional RNN,'' in \textit{2018 21st International Conference on Intelligent Transportation Systems (ITSC)}, Nov 2018, doi: 10.1109/ITSC.2018.8569438.

\bibitem{b5} Y. Xia, D. Zhang, J. Kim, K. Nakayama, K. Zipser, and D. Whitney, ``Predicting Driver Attention in Critical Situations,'' in \textit{Computer Vision – ACCV 2018: 14th Asian Conference on Computer Vision, Perth, Australia, December 2–6, 2018, Revised Selected Papers, Part V}, Springer, 2019, pp. 658--674, doi: 10.1007/978-3-030-20873-8\_42.

\bibitem{b6} P. V. Amadori, T. Fischer, and Y. Demiris, ``HammerDrive: A Task-Aware Driving Visual Attention Model,'' \textit{IEEE Transactions on Intelligent Transportation Systems}, vol. 23, no. 6, pp. 5573--5585, 2022, doi: 10.1109/TITS.2021.3055120.

\bibitem{b7} E. Ohn-Bar and M. M. Trivedi, ``What Makes an On-Road Object Important?'' in \textit{2016 23rd International Conference on Pattern Recognition (ICPR)}, IEEE, 2016, pp. 3392--3397, doi: 10.1109/ICPR.2016.7900158.

\bibitem{b8} M. M. Karim, R. Qin, and Z. Yin, ``An Attention-Guided Multistream Feature Fusion Network for Localization of Risky Objects in Driving Videos,'' 2022, doi: 10.48550/arXiv.2209.07922.

\bibitem{b9} Z. Zhang, A. Tawari, S. Martin, and D. Crandall, ``Interaction Graphs for Object Importance Estimation in On-Road Driving Videos,'' in \textit{2020 IEEE International Conference on Robotics and Automation (ICRA)}, IEEE, 2020, pp. 8920--8927, doi: 10.1109/ICRA40945.2020.9197104.

\bibitem{b10} Z. Zhang, A. Tawari, S. Martin, and D. Crandall, ``Interaction Graphs for Object Importance Estimation in On-Road Driving Videos,'' in \textit{2020 IEEE International Conference on Robotics and Automation (ICRA)}, IEEE, 2020, pp. 8920--8927, doi: 10.1109/ICRA40945.2020.9197104.

\bibitem{b11} Y. Tian, A. Carballo, R. Li, and K. Takeda, ``Road Scene Graph: A Semantic Graph-Based Scene Representation Dataset for Intelligent Vehicles,'' 2020, doi: 10.48550/arXiv.2011.13588.

\bibitem{b12} S.-Y. Yu, A. V. Malawade, D. Muthirayan, P. P. Khargonekar, and M. A. Al Faruque, ``Scene-Graph Augmented Data-Driven Risk Assessment of Autonomous Vehicle Decisions,'' \textit{IEEE Transactions on Intelligent Transportation Systems}, vol. 23, no. 7, pp. 7941--7951, 2022, doi: 10.1109/TITS.2021.3074854.

\bibitem{b13} Y. Niu, M. Ding, K. Fujii, K. Ohtani, A. Carballo, and K. Takeda, ``DRUformer: Enhancing Driving Scene Important Object Detection With Driving Scene Relationship Understanding,'' \textit{IEEE Access}, vol. 12, pp. 67589--67599, 2024, doi: 10.1109/ACCESS.2024.3400589.

\bibitem{b14} J. Li, H. Gang, H. Ma, M. Tomizuka, and C. Choi, ``Important Object Identification With Semi-Supervised Learning for Autonomous Driving,'' in \textit{2022 International Conference on Robotics and Automation (ICRA)}, 2022, pp. 2913--2919, doi: 10.1109/ICRA46639.2022.9812234.

\bibitem{b15} Z. Xu, Y. Zhang, E. Xie, Z. Zhao, Y. Guo, K. K. Wong, Z. Li, and H. Zhao, ``DriveGPT4: Interpretable End-to-End Autonomous Driving Via Large Language Model,'' \textit{IEEE Robotics and Automation Letters}, vol. 9, no. 10, pp. 8186--8193, Oct. 2024, doi: 10.1109/LRA.2024.3440097.

\bibitem{b16} T. Deruyttere, S. Vandenhende, D. Grujicic, L. Van Gool, and M.-F. Moens, ``Talk2Car: Taking Control of Your Self-Driving Car,'' in \textit{Proceedings of the 2019 Conference on Empirical Methods in Natural Language Processing and the 9th International Joint Conference on Natural Language Processing (EMNLP-IJCNLP)}, 2019, pp. 2088--2098, doi: 10.18653/v1/D19-1215.

\bibitem{b17} C. Anderson, R. Vasudevan, and M. Johnson-Roberson, ``Off the Beaten Sidewalk: Pedestrian Prediction in Shared Spaces for Autonomous Vehicles,'' \textit{IEEE Robotics and Automation Letters}, vol. 5, no. 4, pp. 6892--6899, 2020, doi: 10.1109/LRA.2020.3023713.

\bibitem{b18} J. Huang, A. Gautam, and S. Saripalli, ``Learning Pedestrian Actions to Ensure Safe Autonomous Driving,'' in \textit{2023 IEEE Intelligent Vehicles Symposium (IV)}, IEEE, 2023, pp. 1--8, doi: 10.1109/IV55152.2023.10186530.

\bibitem{b19} F. Maresca, F. Grazioli, A. Albanese, V. Sciancalepore, G. Negri, and X. Costa-Perez, ``Are You a Robot? Detecting Autonomous Vehicles from Behavior Analysis,'' in \textit{2024 IEEE International Conference on Robotics and Automation (ICRA)}, IEEE, 2024, pp. 4473--4479, doi: 10.1109/ICRA57147.2024.10610658.

\bibitem{b20} J. Hayakawa and B. Dariush, ``Ego-Motion and Surrounding Vehicle State Estimation Using a Monocular Camera,'' in \textit{2019 IEEE Intelligent Vehicles Symposium (IV)}, IEEE, 2019, pp. 2550--2556, doi: 10.1109/IVS.2019.8814037.

\bibitem{b21} X. Chen, K. Kundu, Z. Zhang, H. Ma, S. Fidler, and R. Urtasun, ``Monocular 3D Object Detection for Autonomous Driving,'' in \textit{2016 IEEE Conference on Computer Vision and Pattern Recognition (CVPR)}, Las Vegas, NV, USA, 2016, pp. 2147--2156, doi: 10.1109/CVPR.2016.236.

\bibitem{b22} A. Mousavian, D. Anguelov, J. Flynn, and J. Kosecka, ``3D Bounding Box Estimation Using Deep Learning and Geometry,'' in \textit{2017 IEEE Conference on Computer Vision and Pattern Recognition (CVPR)}, 2017, doi: 10.1109/CVPR.2017.597.

\bibitem{b23} S. Ren, K. He, R. Girshick, and J. Sun, ``Faster R-CNN: Towards Real-Time Object Detection with Region Proposal Networks,'' \textit{IEEE Transactions on Pattern Analysis and Machine Intelligence}, vol. 39, no. 6, pp. 1137--1149, 2017, doi: 10.1109/TPAMI.2016.2577031.

\bibitem{b24} M. Hussain, ``YOLO-v1 to YOLO-v8: The Rise of YOLO and Its Complementary Nature Toward Digital Manufacturing and Industrial Defect Detection,'' \textit{Machines}, vol. 11, no. 7, p. 677, 2023, doi: 10.3390/machines11070677.

\bibitem{b25} N. Carion, F. Massa, G. Synnaeve, N. Usunier, A. Kirillov, and S. Zagoruyko, ``End-to-End Object Detection with Transformers,'' in \textit{European Conference on Computer Vision (ECCV)}, Springer, 2020, pp. 213--229, doi: 10.1007/978-3-030-58452-8\_13.

\bibitem{b26} R. Birkl, D. Wofk, and M. Müller, ``MiDaS v3.1—A Model Zoo for Robust Monocular Relative Depth Estimation,'' 2023, doi: 10.48550/arXiv.2307.14460.

\bibitem{b27} X. Zhang, R. Abdelfattah, Y. Song, S. A. Dauchert, and X. Wang, ``Depth Monocular Estimation With Attention-Based Encoder-Decoder Network From Single Image,'' in \textit{2022 IEEE 24th International Conference on High Performance Computing \& Communications; 8th International Conference on Data Science \& Systems; 20th International Conference on Smart City; 8th International Conference on Dependability in Sensor, Cloud \& Big Data Systems \& Application (HPCC/DSS/SmartCity/DependSys)}, IEEE, 2022, pp. 1795--1800, doi: 10.1109/HPCC-DSS-SmartCity-DependSys57074.2022.00271.

\bibitem{b28} Y. Hua, Y. Liu, B. Li, and M. Lu, ``Dilated Fully Convolutional Neural Network for Depth Estimation From a Single Image,'' in \textit{2019 International Conference on Computational Science and Computational Intelligence (CSCI)}, IEEE, 2019, pp. 612--616, doi: 10.1109/CSCI49370.2019.00115.

\bibitem{b29} F. Yu, H. Chen, X. Wang, W. Xian, Y. Chen, F. Liu, V. Madhavan, and T. Darrell, ``BDD100K: A Diverse Driving Dataset for Heterogeneous Multitask Learning,'' in \textit{Proceedings of the IEEE/CVF Conference on Computer Vision and Pattern Recognition}, 2020, pp. 2636--2645, doi: 10.1109/CVPR42600.2020.00271.

\bibitem{b30} E. Sachdeva, N. Agarwal, S. Chundi, S. Roelofs, J. Li, M. Kochenderfer, C. Choi, and B. Dariush, ``Rank2Tell: A Multimodal Driving Dataset for Joint Importance Ranking and Reasoning,'' in \textit{Proceedings of the IEEE/CVF Winter Conference on Applications of Computer Vision}, 2024, pp. 7513--7522, doi: 10.1109/WACV57701.2024.00734.

\bibitem{b31} B. Wilson, W. Qi, T. Agarwal, J. Lambert, J. Singh, S. Khandelwal, B. Pan, R. Kumar, A. Hartnett, J. K. Pontes, D. Ramanan, P. Carr, and J. Hays, ``Argoverse 2: Next Generation Datasets for Self-Driving Perception and Forecasting,'' 2023, doi: 10.48550/arXiv.2301.00493.

\bibitem{b32} Y. Xu, X. Yang, L. Gong, H.-C. Lin, T.-Y. Wu, Y. Li, and N. Vasconcelos, ``Explainable Object-Induced Action Decision for Autonomous Vehicles,'' in \textit{Proceedings of the IEEE/CVF Conference on Computer Vision and Pattern Recognition (CVPR)}, 2020, pp. 9520--9529, doi: 10.1109/CVPR42600.2020.00954.

\bibitem{b33} S. Malla, C. Choi, I. Dwivedi, J. H. Choi, and J. Li, ``DRAMA: Joint Risk Localization and Captioning in Driving,'' in \textit{Proceedings of the IEEE/CVF Winter Conference on Applications of Computer Vision}, 2023, pp. 1043--1052, doi: 10.1109/WACV56688.2023.00110.

\bibitem{b34} G. Jocher, J. Qiu, and A. Chaurasia, ``Ultralytics YOLO,'' 2023. [Online]. Available: https://github.com/ultralytics/ultralytics

\bibitem{b35} M. Sandler, A. Howard, M. Zhu, A. Zhmoginov, and L.-C. Chen, ``MobileNetV2: Inverted Residuals and Linear Bottlenecks,'' in \textit{Proceedings of the IEEE Conference on Computer Vision and Pattern Recognition}, 2018, pp. 4510--4520, doi: 10.1109/CVPR.2018.00474.

\bibitem{b36} H. Caesar, V. Bankiti, A. H. Lang, S. Vora, V. E. Liong, Q. Xu, A. Krishnan, Y. Pan, G. Baldan, and O. Beijbom, ``nuScenes: A Multimodal Dataset for Autonomous Driving,'' in \textit{Proceedings of the IEEE/CVF Conference on Computer Vision and Pattern Recognition}, 2020, pp. 11621--11631, doi: 10.1109/CVPR42600.2020.01164.

\bibitem{b37} K. O'Shea and R. Nash, ``An Introduction to Convolutional Neural Networks,'' 2015, doi: 10.48550/arXiv.1511.08458.

\bibitem{b38} A. Vaswani, N. Shazeer, N. Parmar, J. Uszkoreit, L. Jones, A. N. Gomez, Ł. Kaiser, and I. Polosukhin, ``Attention Is All You Need,'' \textit{Advances in Neural Information Processing Systems}, vol. 30, 2017, doi: 10.48550/arXiv.1706.03762.

\bibitem{b39} N. Wojke, A. Bewley, and D. Paulus, ``Simple Online and Realtime Tracking with a Deep Association Metric,'' in \textit{2017 IEEE International Conference on Image Processing (ICIP)}, 2017, pp. 3645--3649, doi: 10.1109/ICIP.2017.8296962.


\end{thebibliography}
\end{document}